\documentclass[11pt,letterpaper]{article}
\usepackage{ijcnlp2017}
\usepackage{times}
\usepackage{latexsym}
\usepackage{amsmath}
\usepackage{amssymb}
\usepackage{multirow}
\usepackage{todonotes}
\usepackage{url}
\usepackage{comment}
\usepackage{algorithmicx}
\usepackage[ruled]{algorithm}
\usepackage[noend]{algpseudocode}
\usepackage[titletoc,toc,title]{appendix}
\usepackage{textcomp}

\newcommand*{\affaddr}[1]{#1} % No op here. Customize it for different styles.
\newcommand*{\affmark}[1][*]{\textsuperscript{#1}}
\newcommand*{\email}[1]{\texttt{#1}}

\usepackage{tikz}
\usepackage{pgfplots}
\usepackage{pgfplotstable}
\usepackage{filecontents}
\usetikzlibrary{pgfplots.groupplots}
\pgfplotsset{compat=1.3}
\usepackage{graphics}
\usetikzlibrary{intersections,positioning}
\usepackage{tikz-dependency}
\usepackage{subcaption}
\usepackage{color}

\ijcnlpfinalcopy

\title{Transferring Semantic Roles Using Translation and Syntactic Information}

\author{%
Maryam Aminian\affmark[1], Mohammad Sadegh Rasooli\affmark[2], Mona Diab\affmark[1]\\
\affaddr{\affmark[1]Department of Computer Science, The George Washington University, Washington}\\
\affaddr{\affmark[2]Department of Computer Science, Columbia University, New York}\\
\affmark[1]\email{\{aminian,mtdiab\}@gwu.edu}\\
\affmark[2]\email{rasooli@cs.columbia.edu}
}

\date{}

\begin{document}

\maketitle

\begin{abstract}
Our paper addresses the problem of annotation projection for semantic role labeling for  resource-poor languages using supervised annotations from a resource-rich language through parallel data. We propose a transfer method that employs information from source and target syntactic dependencies as well as word alignment density to improve the quality of an iterative bootstrapping method. Our experiments yield a $3.5$ absolute labeled F-score improvement over a standard annotation projection method.  
\end{abstract}

\section{Introduction}
\label{sec_intro}

Semantic role labeling (SRL) is the task of automatically labeling  predicates and arguments of a sentence with shallow semantic labels characterizing ``Who did What to Whom, How, When and Where?'' \cite{palmer2010semantic}. These rich semantic representations are useful in many applications such as question answering \cite{shen_srl_qa} and information extraction~\cite{christensen_srl_ie}, hence gaining a lot of attention in recent years \cite{zhou_sup,tackstrom_sup,roth_sup,marcheggiani2017simple}. 
%This illuminates the need for annotated corpora with semantic roles. 
Since the process of creating annotated resources needs significant manual effort, SRL resources are available for a relative small number of languages such as English~\cite{palmer_propbank}, German~\cite{erk_german_srl}, Arabic~\cite{zaghouani_arabic} and Hindi~\cite{vaidya_hindi}. However, most languages still lack SRL systems. There have been some efforts to use information from a resource-rich language to develop SRL systems for resource-poor languages. 
\emph{Transfer} methods address this problem by transferring information from a resource-rich language (e.g. English) to a resource-poor language.
%Besides recent advances in supervised semantic role labeling (e.g. \cite{roth_sup,marcheggiani2017simple}), there have been fewer studies on such systems.

\emph{Annotation projection} is a popular transfer method that transfers supervised annotations from a \emph{source} language to a \emph{target} language through parallel  data.
%Although the scenario for annotation projection seems very straightforward, performance of the SRL system trained on the projected semantic dependencies is not very high.  
Unfortunately this technique is not as straightforward as it seems,
%One of the main causes for this performance degradation are 
e.g. \emph{translation shifts} lead to erroneous projections and accordingly affecting the performance of the SRL system trained on these projections. Translation shifts are typically a result of the differences in word order and the semantic divergences between the source and target languages. %occurs in many cases: this might be from the lack of lexical translation or the difference in the linguistic essence of the source and target languages. 
In addition to translation shifts, there are errors that occur in translations, automatic word alignments as well as automatic semantic roles, hence we observe a cascade of error effect. %Moreover, the annotations in different datasets are heterogeneous, making the evaluation very challenging.   

In this paper, we introduce a new approach for a \emph{dependency-based SRL} system based on annotation projection without any semantically annotated data for a target language. %The annotations from the source language words in each sentence are projected to the target language words by using automatic word alignments. Then, a supervised SRL system trains on the projected data.
%% Maryam  the above sentence is not saying something different from the previous sentence, you are saying yuou are doing the same as the previous approaches. You are still projecting annotations . 
We primarily focus on improving the quality of annotation projection by using  translation cues automatically discovered from word alignments. We show that exclusively relying on partially projected data does not yield good performance. We improve over the baseline by filtering irrelevant projections, iterative bootstrapping with relabeling, and weighting each projection instance differently with data-dependent cost-sensitive training. 

In short, contributions of this paper can be summarized as follows; We introduce a weighting algorithm to improve annotation projection based on cues obtained from syntactic and translation information. In other words, instead of utilizing manually-defined rules to filter projections, we define and use a customized cost function to train over noisy projected instances. This newly defined cost function helps the system weight some projections over other instances. We then utilize this algorithm in a bootstrapping framework. Unlike traditional bootstrapping, ours relabels every training instance (including labeled data) in every self-training round. Our final model on transferring from English to German yields a $3.5$ absolute improvement labeled F-score  over a standard annotation projection method.  

%As a summary of our contributions, we propose a transfer method without using any semantic lexicon and annotated datasets in the target language and achieve improvements by applying cost-sensitive training using syntactic and translation information in a bootstrapping framework. Unlike traditional bootstrapping, our bootstrapping method relabels every training instance (including labeled data) in every self-training round. %: we show a significant improvement by applying this.  We have conducted experiments on transferring information from English to German. %Results show that each of the proposed ideas in this paper gives an improvement in the task. 
%Our final model yields a $3.5$ absolute labeled F-score improvement over a standard annotation projection method.

\section{Our Approach}

We aim to develop a dependency-based SRL system which makes use of training instances projected from a source language (SLang) onto a target language (TLang) through parallel data. %In a dependency-based SRL, for every word in a sentence that is tagged as \emph{predicate}, we have to decide about which words are the \emph{arguments} of that predicate word with what \emph{semantic role}. 
Our SRL system is formed as a pipeline of classifiers consisting of a predicate identification and disambiguation module, an argument identification module, and an argument classification module. In particular, we use our re-implementation of the greedy (local) model of \newcite{bj_sup} except that we use an averaged perceptron algorithm \cite{freund1999large} as the learning algorithm.

\subsection{Baseline Model}\label{sec_baseline}
As our baseline, we apply automatic word alignment on parallel data and preserve the intersected alignments from the source-to-target and target-to-source directions. As our next step, we define a \emph{projection density criteria} to filter some of the projected sentences. Given a target sentence from TLang with $w$ words where $f$ words have alignments ($f\leq w$), if the source sentence from SLang has $p$ predicates for which $p'$ of them are projected ($p'\leq p$), we define projection density as $(p' \times f)/(p \times w)$ and prune out sentences with a density value less than a certain threshold. The threshold value is empirically determined during tuning experiments performed on the development data. In this criteria, the denominator shows the maximum number of training instances that could be obtained by projection and the nominator shows the actual number of relevant instances that are used in our model. In addition to speeding up the training process, filtering sparse alignments helps remove sentence pairs with a significant translation shifts. % (we discuss this criteria more in \S\ref{results_discussion}). 
Thereafter, a supervised model is trained directly on the projected data. 

%%MARYAM here your intuition is that you don't want to train with rarely observed data projections. BUt isn't this subject to the language pair characteristics, some languages would have significant divergences such as Arabic to Chinese due to the relative free word order and morphologically rich nature of the former and the rigid word and lack of morphology for the latter??

\subsection{Model Improvements}

As already mentioned, the quality of projected roles is highly dependent on different factors including translation shifts, errors in automatic word alignments and the SLang supervised SRL system. In order to address these problems, we apply the following techniques to improve learning from partial and noisy projections, thereby enhancing the performance of our model: 
\begin{itemize}
    \item Bootstrapping to make use of unlabeled data;
    \item Determining the quality of a particular projected semantic dependency based on two factors: 1) source-target syntactic correspondence; and, 2) projection completeness degree. We utilize the above constraints in the form of a data-dependent cost-sensitive training objective. This way the classifier would be able to learn translation shifts and erroneous instances in the projected data, hence enhancing the overall performance of the system. 
\end{itemize}
\paragraph{Bootstrapping}
Bootstrapping (or self-training) is a simple but very useful technique that makes use of unlabeled data. A traditional self-training method \cite{mcclosky_self} labels unlabeled data (in our case, fill in missing SRL decisions) and adds that data to the labeled data for further training. We report results for this setting in \S\ref{results_discussion} as \emph{fill--in}. Although fill--in method is shown to be very useful in previous work \cite{akbik_projection}, empirically, we find that it is better to \emph{relabel} \textbf{all} training instances (including the already labeled data) instead of only labeling unlabeled raw data. Therefore, the classifier is empowered to discover outliers (resulting from erroneous projections) and change their labels during the training process. Figure~\ref{fig_bootstrap} illustrates our algorithm. It starts with training on the labeled data and uses the trained model to label the unlabeled data and relabel the already labeled data. This process repeats for a certain number of epochs until the model converges, i.e., reaches its maximum performance. 
%We provide two modes for bootstrapping: in the default mode, we just relabel the data that was originally unlabeled (\emph{supplement} in Figure\ref{fig_bootstrap}); and the other mode is to relabel all training instances (\emph{relabel} in Figure\ref{fig_bootstrap}). In our experiments we found out that the \emph{supplement} approach does slightly better so we stick with it throughout the paper. 

%%MARYAM in this above paragraph you seem to indicate that supplement does better than relabel ... which of them actually works better? I think it might be worth presenting both? with numbers --> %%MONA: I reported the results for both cases in the results section. I also have an example that shows how different iterations of relabeling helps refine labels but space won't allow to add that example.
\input{bootstrap}
\paragraph{Data-dependent cost-sensitive training}

In our baseline approach, we use the standard perceptron training. In other words, whenever the algorithm sees a training instance $x_i$ with its corresponding label $y_i$, it updates the weight vector $\theta$ for iteration $t$ based on the difference between the feature vector $\phi(x_i,y_i)$ of the gold label and the feature vector $\phi(x_i,y_i^*)$ of the predicted label $y_i^*$ (Eq.~\ref{eq:perceptron_update}).
\begin{equation}
\theta^{t} = \theta^{t-1} + \phi(x_i, y_i) - \phi(x_i, y_i^*) 
\label{eq:perceptron_update}
\end{equation}

In Eq.~\ref{eq:perceptron_update}, the algorithm assumes that every data point $x_i$ in the training data $\{x_1, \cdots, x_n\}$ has the same importance and the cost of wrongly predicting the best label for each training instance is uniform. We believe this uniform update is problematic especially for the transfer task in which different projected instances have different qualities. To mitigate this issue, we propose a simple modification, we introduce a cost $\lambda_i \in [0,1]$ for each training instance $x_i$. Therefore, Eq.~\ref{eq:perceptron_update} is modified as follows in Eq.~\ref{eq:modified_perceptron_update}.
\begin{equation}
\theta^{t} = \theta^{t-1} + \lambda_i ~ ( \phi(x_i, y_i) - \phi(x_i, y_i^*)) 
\label{eq:modified_perceptron_update}
\end{equation}

In other words, the penalty of making a mistake by the classifier for each training instance depends on the importance of that instance defined by a certain cost. The main challenge is to define an effective cost function, especially in our framework where we don't have supervision. Accordingly, we experiment with the following cost definitions:
\begin{itemize}
    \item[-]{\bf Projection completeness}: Our observation shows that the density of projection is a very important indicator of projection quality. We view it as a rough indicator of translation shifts: the more alignments
    %%MARYAM what do you mean by decisions, you mean alignments??-->%%MONA: addressed
    from  source to  target, the less we have a chance of having translation shifts. 
    As an example, consider the sentence pair extracted from English--German Europarl corpus: ``\emph{I sit here between a rock and a hard place}'' and its German translation ``\emph{Zwei Herzen wohnen ach in meiner Brust}'' which literally reads as ``\emph{Two hearts dwell in my chest}''. The only words that are aligned (based on the output of Giza++) are the English word ``\emph{between}'' and the German word ``\emph{in}''. In fact, the German sentence is an idiomatic translation of the English sentence. Consequently predicate--argument structure of these sentences vary tremendously; The word ``\emph{sit}'' is predicate of the English sentence while ``\emph{wohnen} (dwell)'' is the predicate of the German sentence.

    We use the definition of \emph{completeness} from \newcite{akbik_projection} to define the sparsity cost ($\lambda^{\small{comp}}$): this definition deals with the proportion of a verb or direct dependents of verbs in a sentence that are labeled. 
    %%MARYAM I think you need to provide a example here --> %%MONA: I can give more details using sentences in Fig.2 but I skipped the details here to save the space for the above example.
    
    \item[-]{\bf Source-target syntactic dependency match}: %Like previous work \cite{puny_syn_sem}, we find that syntactic information is a valuable clue in SRL. 
    We observe that when the dependency label of a target word is different from its aligned source word, there is a higher chance of a projection mistake. However, given the high frequency of source-target dependency mismatches, it is harmful to prune those projections that have dependency mismatch; instead, we define a different cost if we see a training instance with a dependency mismatch. For an argument $x_i$ that is projected from source argument $s_{x_i}$, we define the cost $\lambda^{\small{dep}}_i$ according to the dependency of the source and target words $dep(x_i)$ and $dep(s_{x_i})$ as Eq.~\ref{eq:dep_match}.
    %%MARYAM I think you need to provide a example here-->%%MONA: addressed
    \begin{equation}
    \lambda_i^{dep} =
    \begin{cases}
    1 & \text{if}~dep(x_i) = dep(s_{x_i}) \\
    0.5 &  \text{otherwise} \\
    \end{cases}
    \label{eq:dep_match}
    \end{equation}
    As an example, consider Fig.~\ref{fig:dep_match} that demonstrates an English-German sentence pair from EuroParl ``\emph {I would urge you to endorse this}'' with its German translation that literally reads as ``\emph{I ask for your approval}''. As we can see, there is a shift in translation of English clausal complement ``\emph{to endorse this}'' into German equivalent ``\emph{um Zustimmung (your approaval)}'' which leads the difference in the syntactic structure of source and target sentences. Therefore, neither the predicate label of English verb ``\emph{endorse}'' nor the argument ``\emph{A2}'' should not be projected to the German noun ``\emph{Zustimmung}''. Dashed edges between sentences show intersected word alignments. Here, projecting semantic role of ''\emph{endorse}`` (A2) to to the word ''\emph{Zustimmmung}`` through alignment will lead to the wrong semantic role for this word.

    %Figure~\ref{fig:dep_match_w_srl} shows the output of annotation projection from English to German which leads to the wrong semantic role ``A2'' for argument ``Zustimmmung''. Figure~\ref{error_anal_example_b} shows the output of the first relabeling iteration. As we can see, applying the source--target dependency match constraint for training helps our system identify the translation shift and filter out argument ``\emph{A2}''.
    
    \begin{figure}[!t]
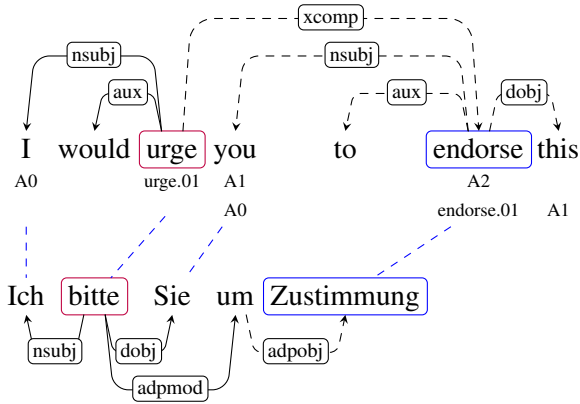


%\vspace{-1cm} 
\centering
\begin{dependency}%[theme = simple]
\begin{deptext}
I\&would\&urge\&you\&to\&endorse\&this\\
\scriptsize{A0}\&\&\scriptsize{urge.01}\&\scriptsize{A1}\&\&\scriptsize{A2}\&\\
\&\&\&\scriptsize{A0}\&\&\scriptsize{endorse.01}\&\scriptsize{A1}\\
\\
\\
Ich\&bitte\&Sie\&um\&Zustimmung\&\\
\end{deptext}

\depedge{3}{1}{nsubj}
\depedge{3}{2}{aux}
\depedge[dashed]{6}{4}{nsubj}
\depedge[dashed]{6}{5}{aux}
\depedge[dashed]{3}{6}{xcomp}
\depedge[dashed]{6}{7}{dobj}

\depedge[edge below]{2}{1}{nsubj}
\depedge[edge below]{2}{3}{dobj}
\depedge[edge below]{2}{4}{adpmod}
\depedge[dashed,edge below]{4}{5}{adpobj}

\path[dashed,color=blue] (\wordref{3}{1}) edge (\wordref{6}{1});
\path[dashed,color=blue] (\wordref{3}{3}) edge (\wordref{6}{2});
\path[dashed,color=blue] (\wordref{3}{4}) edge (\wordref{6}{3});
\path[dashed,color=blue] (\wordref{3}{6}) edge (\wordref{6}{5});

\wordgroup[color=purple]{1}{3}{3}{pred}
\wordgroup[color=blue]{1}{6}{6}{pred}

\wordgroup[color=purple]{6}{2}{2}{pred}
\wordgroup[color=blue]{6}{5}{5}{pred}

\end{dependency}

\caption{Example of English-German sentences from Europarl with dependency structure. Different dependencies are shown with dashed arcs. Predicate--argument structure of the English sentence is shown bellow each word.}\label{fig:dep_match}
\end{figure}

    \item[-] {\bf Completeness + syntactic match}: We employ the average of $\lambda^{\small{dep}}$ and $\lambda^{\small{comp}}$ values as defined above. This way, we simultaneously encode both the completeness and syntactic similarity information.
\end{itemize}

\section{Experiments}
\label{sec_experiments}

\paragraph{Data and Setting}
We use English as the source and German as the target language. In our setting, we assume to have supervised part-of-speech tagging and dependency parsing models for both the source (SLang) and target (TLang)  languages. We use the Universal part-of-speech tagset of \newcite{petrov2011universal} and the Google Universal Treebank \cite{google_univ}. We ignore the projection of the \emph{AM} roles to German since this particular role does not appear in the German dataset.

We use the standard data splits in the CoNLL shared task on SRL \cite{conll2009_shared} for evaluation. We replace the POS and dependency information with the predictions from the Yara parser \cite{rasooli2015yara} trained on the Google Universal Treebank.\footnote{Our ideal setting is to  transfer to more languages but because of the semantic label inconsistency across CoNLL datasets, we find it impossible to evaluate our model on more languages. Future work should work on defining a reliable conversion scheme to unify the annotations in different datasets.} We use the parallel Europarl corpus \cite{koehn2005europarl}  and Giza++ \cite{och2000giza} for extracting word alignments. Since predicate senses are projected from English to German, comparing projected senses with the gold German predicate sense is impossible. To address this, all evaluations are conducted using the Gold predicate sense.

After filtering projections with density criteria of \S\ref{sec_baseline}, 29417 of the sentences are preserved. The number of preserved sentences after filtering sparse alignments is roughly one percent of the original parallel data (29K sentences out of 2.2M sentences). Density threshold is set to $0.4$ determined based on our tuning experiments on development data.  

\subsection{Results and Discussion}
\label{results_discussion}
Table \ref{tab_de_results} shows the results of different models on the German evaluation data. As we can see in the table, bootstrapping outperforms the baseline. Interestingly, relabelling all training instances (Bootstrap--relabel) gives us $0.8$ absolute improvement in F-score compared to when we just predict over instances without a projected label (Bootstrap--fill-in). Here, the fill-in approach would label only the German word ''\emph{um}`` in Fig.~\ref{fig:dep_match} that does not have any projected label from the English side. While the relabeling method will overwrite all projected labels with less noisy predicted labels.
%%MARYAM WHat is this fill in case, please explain it and give an example  

\begin{figure}
\centering
\begin{tikzpicture}
\pgfplotsset{small,samples=20}
    \begin{groupplot}[group style = {group size = 2 by 1, horizontal sep = 30pt}, width = 4.0cm, height = 3.8cm]
    \nextgroupplot[title ={\scriptsize{VERB+A0}}, legend style = { column sep = 7pt, legend columns = 3, legend to name = grouplegend,}]
            \addplot[
                color=blue,
                mark=square,
                ]
                coordinates {
                (1,0.6090)(2,0.6093)(3,0.6023)(4,0.6145)(5,0.6102)(6,0.6130)(7,0.6218)
                };
                \addlegendentry{\small{precision}}%
            
                \addplot[
                    color=red,
                    mark=triangle,
                    ]
                    coordinates {
                    (1,0.4904)(2,0.5024)(3,0.5024)(4,0.5288)(5,0.5192)(6,0.5216)(7,0.5337)
                    };
                    \addlegendentry{\small{recall}}

                \addplot[
                    color=green,
                    mark=star,
                    ]
                    coordinates {
                    (1,0.5433)(2,0.5507)(3,0.5478)(4,0.5684)(5,0.5610)(6,0.5636)(7,0.5744)
                    };
                    \addlegendentry{\small{f-score}}
    \nextgroupplot[title = {\scriptsize{VERB+A1}},]
             
            \addplot[
                color=blue,
                mark=square,
                ]
                coordinates {
                (1,0.4829)(2,0.4743)(3,0.4929)(4,0.5103)(5,0.5088)(6,0.5088)(7,0.5146)
                };
                
            \addplot[
                color=red,
                mark=triangle,
                ]
                coordinates {
                (1,0.4340)(2,0.4127)(3,0.4104)(4,0.4104)(5,0.4080)(6,0.4080)(7,0.4151)
                };
                
            \addplot[
                color=green,
                mark=star,
                ]
                coordinates {
                (1,0.4571)(2,0.4414)(3,0.4479)(4,0.4549)(5,0.4529)(6,0.4529)(7,0.4595)
                };

    \end{groupplot}
    \node at ($(group c2r1) + (-1.5,-1.85cm)$) {\ref{grouplegend}}; 
\end{tikzpicture}
\caption{Precision, recall and F--score of VERB+A0 and VERB+A1 during relabeling iterations on the German development data. Horizontal axis shows the number of iterations and vertical axis shows values of precision, recall and F--score.}\label{precision_recall}
\end{figure}
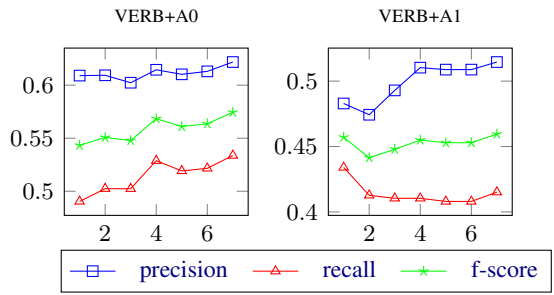

We additionally observe that the combination of the two cost functions improves the quality further. 
%\section{Discussion}
Overall, the best model yields $3.5$ absolute improvement F-score  over the baseline. As expected, none of the approaches improves over supervised performance.  
\begin{table}[t!]
    \centering
   %\small
    \begin{tabular}{l| c  c}
    \hline \hline
      Model   & Cost & Lab. F1  \\ \hline \hline
      Baseline &$\times$ & 60.3 \\ \hline
   Bootstrap--fill-in &   $\times$ &  61.6 \\
Bootstrap--relabel &  $\times$  & 62.4 \\ \hline
 Bootstrap--relabel &   comp. & 63.0{\footnotesize$_{(+1.0)}$} \\
     Bootstrap--relabel &   dep. &  63.4{\footnotesize$_{(+1.8)}$} \\ 
       Bootstrap--relabel &   comp.+dep.  & {\bf 63.8}{\footnotesize$_{(+1.3)}$} \\ \hline
    Supervised & -- &  79.5 \\ \hline \hline
    \end{tabular}
    \caption{Labeled F-score for different models in SRL transfer from English to German  using gold predicates. Cost columns shows the use of cost-sensitive training using projection completeness (``comp.''), source-target dependency match (``dep.'') and both (``comp.+dep.''). The numbers in parenthesis show the absolute improvement over the Bootstrap-fill-in method.}
    \label{tab_de_results}
\end{table}
We further analyzed the effects of relabeling approach on identification and classification of non--root semantic dependencies. Figure~\ref{precision_recall} shows precision, recall and F--score of the two most frequent semantic dependencies (predicate pos + argument label): VERB+A0, VERB+A1 throughout relabeling iterations. As demonstrated in the graph, both precision and recall improve by cost-sensitive relabeling for VERB+A0. In fact, cost-sensitive training helps the system refine irrelevant projections at each iteration and assigns more weight on less noisy projections, hence enhancing precision. Our analysis on VERB+A0 instances shows that source--target dependency match percentage also increases during iterations leading to increase the recall. In other words, weighting projection instances based on dependency match helps classifier label some of the instances which were dismissed during projection, thereby will increase the recall. While similar improvement in precision is observed for VERB+A1, Figure~\ref{precision_recall} shows that the recall is almost descending by relabeling. Our analysis shows that unlike VERB+A0, percentage of source--target dependency match remains almost steady for VERB+A1. This means that cost-sensitive relabeling for this particular semantic dependency has not been very successful in labeling unlabeled data.

\section{Related Work}

%While the majority of work on SRL has focused on supervised learning, 
There have been several studies on transferring SRL systems~\cite{pado2005cross,pado2009cross,mukund_projection,vanderplas_cross,vanderplas_cross_global,kozhevnikov_transfer,akbik_projection}. \newcite{pado2005cross}, as one of the earliest studies on annotation projection for SRL using parallel resources, apply different heuristics and techniques to improve the quality of their model by focusing on having better word and constituent alignments. \newcite{vanderplas_cross} improve an annotation projection model by jointly training a transfer system for parsing and SRL. They solely focus on fully projected annotations and train only on verbs. In this work, we train on all predicates as well as exploit partial annotation. \newcite{kozhevnikov_transfer} define shared feature representations between the source and target languages in annotation projection. The benefit of using shared representations is complementary to our work encouraging us to use it in future work. 

\newcite{akbik_projection} introduce an iterative self-training approach using different types of linguistic heuristics and alignment filters to improve the quality of projected roles. Unlike our work that does not use any external resources, \newcite{akbik_projection} make use of bilingual dictionaries. Our work also leverages self-training but with a different approach: first of all, ours does not apply any heuristics to filter out projections. Second, it trains and relabels  all projected instances, either labeled or unlabeled, at every epoch and does not gradually introduce new unlabeled data. Instead, we find it more useful to let the target language SRL system rule out noisy projections via relabeling. %It is worth noting that there are many studies on improving supervised SRL using parallel data (e.g. \cite{titov_crosslingual,kozhev_bootstrap}).

%\newcite{pascale_biframe} sfsf

%\newcite{kozhev_bootstrap} sasd 

%\cite{kozhev_bootstrap,pascale_biframe,mukund_projection,vanderplas_cross}

%\cite{pado_projection,pado2005cross,akbik_projection,kozhevnikov_transfer,titov_crosslingual,vanderplas_cross_global} 

%\cite{lang_unsup,garg_unsup,titov_unsup,luan_unsup,lang_unsup_graph,lorenzo_unsup,lang_unsup_split,khoddam_unsup,lewis_unsup}

%\cite{titov_semi,hagen_semi}

\section{Conclusion}

We described a method to improve the performance of annotation projection in the dependency-based SRL task utilizing a data-dependent cost-sensitive training. Unlinke previous studies that use manually-defined rules to filter projections, we benefit from information obtained from projection sparsity and syntactic similarity to weigh projections. We utilize a bootstrapping algorithm to train a SRL system over projections. We showed that we can get better results if we relabel the entire train data in each iteration as opposed to only labeling instances without projections.    

%Due to different annotation schemes for different languages in currently available standard data sets (e.g. CoNLL data set), we can not perform a meaningful comparison with other studies. 

For the future work, we consider experimenting with newly published Universal Proposition Bank~\cite{universal_prop_bank} that provides a unified labeling scheme for all languages. Given the recent success in SRL systems with neural networks \cite{marcheggiani2017simple,marcheggiani2017encoding}, we plan to use them for further improvement. We expect a similar trend by applying the same ideas in a neural SRL system.

\section*{Acknowledgments}
This work has been partly funded by DARPA LORELEI Grant and generous support by Leidos Corp. for the 1st and 3rd authors. We would like to acknowledge the useful comments by three anonymous reviewers who helped in making this publication more concise and better presented.

\bibliography{ijcnlp2017}
\bibliographystyle{ijcnlp2017}

\end{document}